\documentclass{bmcart}

\usepackage[utf8]{inputenc} 
\usepackage{graphicx}

\usepackage{longtable}
\usepackage{amsfonts}

\startlocaldefs
\endlocaldefs

\begin{document}

\begin{frontmatter}

\begin{fmbox}
\dochead{Research}


\title{A summary of the prevalence of Genetic Algorithms in Bioinformatics from 2015 onwards}


\author[
   addressref={aff1},                   
   noteref={n1},                        
   email={mdswerhu@lakeheadu.ca}   
]{\inits{MS}\fnm{Mekaal} \snm{Swerhun}}
\author[
   addressref={aff1},
   noteref={n1},
   email={jkfoley@lakeheadu.ca}
]{\inits{JF}\fnm{Jasmine} \snm{Foley}}
\author[
   addressref={aff1},
   noteref={n1},
   email={bdmossop@lakeheadu.ca}
]{\inits{BM}\fnm{Brandon} \snm{Mossop}}
\author[
   addressref={aff1},
   email={vmago@lakeheadu.ca}
]{\inits{VM}\fnm{Vijay} \snm{Mago}}

\address[id=aff1]{
  \orgname{Department of Computer Science, Lakehead University}, 
  \street{955 Oliver Rd},                     %
  \postcode{P7B 5E1}                        
  \city{Thunder Bay},                              
  \cny{Canada}                                    
}


\begin{artnotes}
\note[id=n1]{Equal contributor} 
\end{artnotes}

\end{fmbox}


\begin{abstractbox}

\begin{abstract} 
\parttitle{Background}
In recent years, machine learning has seen an increasing presence in a large variety of fields, especially in health care and bioinformatics. More specifically, the field where machine learning algorithms have found most application is Genetic Algorithms. 

\parttitle{Objective}
The objective of this paper is to conduct a survey of articles published from 2015 onwards that deal with Genetic Algorithms and how they are used in bioinformatics.  

\parttitle{Methods}
To achieve the objective, a scoping review was conducted that  utilized Google Scholar alongside Publish or Perish and the Scimago Journal\& Country Rank to search for respectable sources. 

\parttitle{Results}
Upon analyzing 31 articles from the field of bioinformatics, it became apparent that genetic algorithms rarely form a full application, instead they rely on other vital algorithms such as support vector machines. Indeed, support vector machines were the most prevalent algorithms used alongside genetic algorithms (GA); however, while the usage of such algorithms contributes to the heavy focus on accuracy by GA programs, it often sidelines computation times in the process. In fact, most applications employing GAs for classification and feature selection are nearing or at 100\% success rate, and the focus of future GA development should be directed elsewhere.

\parttitle{Conclusion}
Population-based searches, like GA, are often combined with other machine learning algorithms. In this scoping review, genetic algorithms combined with Support Vector Machines were found to perform best. The performance metric that was evaluated most often was the accuracy. Measuring the accuracy avoids measuring the main weakness of GAs, which is computational time. The future of genetic algorithms could be “open-ended” evolutionary algorithms, which attempt to increase complexity and find diverse solutions, rather than optimize a fitness function and converge to a single “best” solution from the initial population of solutions.

\end{abstract}




\end{abstractbox}
%

\end{frontmatter}



\section*{1 Introduction}

\subsection*{Genetic Algorithms}
Genetic Algorithms (GA) belong to a larger class of evolutionary algorithms.  A parallel search heuristic algorithm inspired by Charles Darwin's theory of natural selection is modeled by the guiding principle of \textit{Survival of the Fittest} \cite{lu2017hybrid}.  The algorithm selects the fittest individuals of the population with the aim of producing offspring for the next generation that inherit the optimal characteristics of the parents.  This process continues to iterate developing sequential populations, until it converges on a generation with the fittest individuals \cite{alivckovic2017breast}.  Similarly, GA solves problems by optimizing a single criterion, known as a fitness function.  The fitness function estimates the importance by assigning a value to each chromosome that relates to its ability to solving the problem \cite{alivckovic2017breast, salem2017classification}.  A chromosome could be an array of numbers, a binary string, or a list of instances in a database, all relating to and depending on the problem. Each individual that forms the population, represents different possible solutions. Chromosomes deemed fitter have an increased likelihood of being used in the following generation.  The individuals proceed through a process of evolution, which are is of the principles of mutation, selection, and crossover all impacting the fitness value \cite{alivckovic2017breast, subasi2019epileptic}. The most noteworthy benefit about GA is its ability to search sophisticated and massive spaces proficiently and identify near optimal solutions rapidly \cite{salem2017classification}.  Often in order to achieve better performance, GA-based selected features are applied as input to classifiers \cite{alivckovic2017breast}.

\subsection*{Popularity of Genetic Algorithms in Biomedical Applications}
While the properties accredited to GAs make them desirable to a variety of fields, their use in biomedical applications is far-ranging and well-established as shall be made evident in this article. In the medical field GA-based solutions have been posed for a variety of problems including symptom and ailment classification \cite{salem2017classification, subasi2019epileptic, zhang2019selection}, visualization \cite{mohammed2018real} as well as identification and diagnoses of diseases \cite{alivckovic2017breast,sayed2019nested}. GA-based solutions have also increasingly been used at the molecular level in tasks such as handling and predicting transposon-derived piRNAs \cite{li2016genetic}. Yet the importance of GA-based solutions in the medical field is not limited to solving  problems on the microscopic scope as applications have been developed to handle larger scale infrastructure and logistics that can be vital for entire health care systems \cite{khanduzi2019fast, fogue2016non}. Among the most frequent uses of GAs however, is their role in feature selection where they help to narrow down the possible features so that a complementary algorithm can achieve far greater performance \cite{sayed2019nested, soufan2015dwfs, oztekin2018decision, plawiak2018novel,mitchell1998introduction}. At times, a GA-based solution may involve the GA fill multiple of the above mentioned roles such as finding usage in both feature selection and classification. Of course, other GA applications beyond what has already been mentioned exist; however, the applications mentioned here show just how important GA has become in the biomedical field and these are the most common uses found in the papers surveyed in this article.       

\subsection*{Key Findings of the Survey}
While conducting research, a few key points have been discerned which frequently appeared in the selected papers for this survey article. These have been summarized below.
\begin{itemize}
    \item Applications often use GA alongside other machine learning algorithms, most commonly classification algorithms.
    \item Among classification engines used in conjunction with GA, Support Vector Machines (SVM) is the top performing.
    \item Accuracy is one of the prime evaluation metrics focused on; while computation time is often ignored or under-performing for usage in live biomedical situations.
    \item In general, applications employing GAs for classification, and feature selection are reaching close to perfect and at times even perfect results.
\end{itemize}
\subsection*{Structure of the Paper} The following sections of this survey article are organized as follows: Section 2 focuses on the thirty-one papers surveyed for this article. This section first discusses the methodology explaining how the papers were selected before discussing the biomedical issues the papers investigate. Section 2 concludes with a discussion of the common data sets and tools used within the papers. In section 3, the focus is on how the researchers evaluate their studies, with the various performance metrics used being examined and explained to discern the advantages and disadvantages of prioritizing one metric over another. Next in section 4, this article briefly discusses the future of GA. The final section concisely concludes the findings of this survey article.
\section*{2 Methodology}
\subsection*{Paper Selection}
The proposed searching procedure in this survey aims to outline a simple yet effective sequence of operations in order to identify and select high quality manuscripts published in journals.  While utilizing Google Scholar and/or Publish or Perish \cite{publishORperish}, the first step was to establish the date range of the journals published starting with 2015 and proceeding onward.  This survey focuses on the applications of GA, which yields a wide range of possibilities. Therefore, in order to narrow the scope, additional key search terms were needed. In step two, additional key terms, such as biomedical/medicine, and machine learning, were used alongside the main search term. Once a paper was identified it was added to a list of prospective sources. The quality of the paper was examined and identified in step three by utilizing Scimago Journal \& Country Rank (SJR)\cite{sjr} to access the quality of the journal where the paper has been published.  Papers published in journals with a journal ranking of Q2, Q3, and Q4 were immediately removed from the list, and papers published in journals with a journal ranking of Q1 at time of publication were kept.  Once the paper met the quality criteria for its journal ranking, step four ensured if the GA has a dominant role or is used as a key element in the paper.  If the paper does not have either, it was removed from the list.  Papers that had GA serving a dominant role or where GA was used as a key element were kept, further analyzed, and contributed to this survey.  Therefore, each paper had to meet all of the above requirements set in place to be selected.  The whole process is illustrated and can be found as a flowchart in Figure ~\ref{fig:search_diagram}. As a result of this searching methodology, a total of 31 papers were selected for this survey and can be found in Table~\ref{tab:papers}. 

\subsection*{Applications of GA in Bioinformatics}
Using the described searching procedure above, Table~\ref{tab:papers} provides a summary containing key information on the papers selected for this survey. In addition to Table~\ref{tab:papers}, Table~\ref{tab:reproduce} shows the extent to which results could be replicated to obtain similar findings to those papers studied in this survey.  Yet Table~\ref{tab:reproduce} also serves to highlight a concerning issue as it shows how few papers provide the necessary information needed in order for others to reproduce their results. All chosen papers discuss the possible and proposed application in biomedical applications, and are limited to SJR Q1 rankings, from 2015 and later.  Key findings included in Table~\ref{tab:papers} in addition to the biomedical application were examined, the use of GA was noted, and the benefits of the proposed application were identified. 19 of the 31 papers surveyed mention the GA playing a key role in feature selection. Feature selection is a data pre-processing technique that reduces the overall number of features by eliminating redundant samples \cite{lu2017hybrid}.  The task of feature selection is to extract those features that are deemed the most informative and important in predicting the outcome for an individual \cite{alivckovic2017breast}. This technique is an essential step in reducing the dimensionality of the search space and the computational complexity.  Alongside feature selection, GAs are commonly used in classification programs.  Just about half of the papers surveyed, 16 out of the 31, use a form of classification.  Classification aims to predict outcomes associated with a particular individual given a feature vector describing that individual.  GA provides an efficient and robust feature selection algorithm that speeds up the learning process of classifiers and stabilizes the classification accuracy. Within bioinformatics, feature selection and classification both serve vital roles and can often be found within the same program, with the GA selecting features that are then used by a separate algorithm to assign a label that may be a diagnosis of a general disease or even the identification of symptoms. In recent years, GA-based applications have developed to not only identify ailments, but recommend what treatments should be used to combat an ailment that has appeared in different patients \cite{zhang2019selection}. GA also has been utilized in non-standard implementations such as running multiple GA in parallel \cite{soufan2015dwfs}, or nested inside one another as in \cite{sayed2019nested}, which has allowed for the diagnosis and identification of different cancers biomarkers. Indeed, non-standard implementations have even allowed for a hybrid GA-based application to be created that can determine the person to receive the highest quality of life improvement from a lung transplant, helping to ensure that any unforeseen bias does not effect the transplant \cite{oztekin2018decision}. Additionally, GAs have been used for imaging and visualizing applications both due to their importance in feature selection and their ability to combine representations of learned information such as known shapes, and relative position into a single framework that can be used in three-dimensional segmentation  \cite{ghosh2016incorporating}. Finally GAs have been employed to handle logistics both in handling complex hospital supply chains \cite{khanduzi2019fast} and in optimizing ambulance dispatches to non-emergency situations \cite{fogue2016non}. Therefore, it can be easily seen that bioinformatics research entails many problems that can be solved using machine learning tasks, and that GA is well-suited for such tasks.  Yet, it is important that research conducted in this area be highly accurate, efficient, and reliable in order for the results to be meaningful. They need to be prompt and able to withstand the volatile situations that can be found in this field, especially since such solutions are becoming prevalent in nearly every aspect of bioinformatics.  

\subsection*{Datasets}
In order to learn more about how the papers selected for this survey came to their conclusions, a closer look was given to the data used and the sources of the data. Out of the 31 surveyed papers, not a single one used the exact same raw data. Three general patterns emerge from the diversity of datasets.

The most common method for data acquisition in the 31 papers was conducted by accessing digital repositories to find datasets relevant for the topic of the paper.  These repositories act as a tool, compiling datasets that are available to the public and therefore allowing researchers to focus on their project immediately rather than having to conduct a multitude of tests just to acquire data to use for testing. Some examples of repositories seen in the surveyed papers are as follows.
\begin{itemize}
    \item UCSC Genome Browser used by both Li et al. (2016) and Tangherloni et al. (2019) provides access to assembled genomes including the human genome \cite{li2016genetic, tangherloni2019genhap}.
    \item Gene Expression Omnibus used by Sayed et al. (2019) provides more specialized data related to genomics and is itself part of the National Center for Biotechnology Information data resources \cite{sayed2019nested}. 
    \item Protein Data Bank used by Moraes et al. (2017) provides data relating to wide selection of proteins and related components \cite{moraes2017gass}.
\end{itemize}
Besides acquiring data from public repositories, another method of data acquisition employed by some of the surveyed papers was requesting access to data that is generally kept private. Among the sources for this type of data, private databases curated by institutions were the most common. It is important to note that not all required a paper’s authors to be a member of the institution as is the case in Oztekin et al. (2018), who accessed their data from the United Network for Organ Sharing \cite{oztekin2018decision}. In addition, some data sources originate from entities whose primary concern was not data curation, but who could grant access to records of their regular functions. One instance of such data collection can be seen in the work of Fogue et al. (2016) who received their data from an Ambulance Company based in Husca, Spain \cite{fogue2016non}. The final method of data acquisition employed was only used by a minority of the papers surveyed -creation of the data by the project members \cite{liu2019deep}. This final method although being necessary in cases where the data needed is not available does not ensure an unbiased result and would consume significant time for properly compiling the information. Indeed, it would appear to be that due to these downsides, this method of data acquisition is far from favoured. 

Despite the prevalence of acquiring data from pre-existing sources, the raw data acquired often has to go through preprocessing before it is used. What this entails can be widely different depending on the source of the data and its intended purpose; however, most commonly the goal is to narrow down the raw data into a set deemed usable for the project. Such a process may be necessary because in some cases a) the raw dataset does not have enough records, or b) not all records are complete, or c) records are not usable (too much noise) \cite{plawiak2018novel}. A summary of the datasets used by the 31 surveyed papers and their sources can be found in Table~\ref{tab:data}.

\subsection*{Tools}
In addition to looking at what datasets the surveyed papers use, this paper takes a look at the tools and additional machine learning algorithms employed alongside the GA, although a few papers rely solely on GA. Indeed, when looking at the surveyed papers it would appear that GA-focused solutions benefit the most when they are supported by complimentary tools and algorithms. The use of components is much like the datasets mentioned, where a wide variety was used in each study to achieve the goal of that particular paper. However, unlike the datasets a few tools and additional machine learning algorithms were employed across multiple papers fairly regularly. The full selection of tools and machine learning algorithms employed has been compiled in Table~\ref{tab:tools}.

Amongst the 31 surveyed articles, two tools proved to be the most prevalent. The first of these is MATLAB, which is used in \cite{mohammed2018real,khanduzi2019fast,soufan2015dwfs, hashem2017comparison, ghosh2016incorporating, hemanth2019modified, tan2016genetic, plawiak2018novel}. The second tool is Weka, which sees usage in \cite{al2017examining, alivckovic2017breast, gangavarapu2019novel, hashem2017comparison}. MATLAB is a fairly well-known and important tool in studies such as signal processing, data analytics, image processing, and machine learning, partially due to its versatility. In fact, even though all the surveyed papers have a focus on GAs, the way that MATLAB is utilized varies from paper to paper. For instance Soufan et al. (2015) only makes limited use of MATLAB to ensure fairness when evaluating programs \cite{soufan2015dwfs}. P{\l}awiak (2018) uses MATLAB alongside the library, LIBSVM, to implement their study \cite{plawiak2018novel}. 

Weka is a more specialized tool that provides an environment for classification, regression, clustering, and feature selection. It accomplishes this by aiding its users in the extraction of information and helping them find suitable algorithms for creating accurate predictive model with that information \cite{frank2004data}. Although Weka has a far smaller toolbox, it can be ideal for researchers working in bioinformatics due to its focus. Indeed, both of these tools have proven beneficial for a number of the surveyed articles as shown by Hashem et al. (2017) who use both tools to perform algorithms such as Particle Swarm Optimization \cite{hashem2017comparison}. 

Throughout the surveyed articles, additional machine learning algorithms are often used alongside the GA, where they prevalently serve as classification algorithms. The goal of such algorithms is to be able to predict successfully the correct outcome that is associated with a particular occurrence after having received a selection of features that describe the occurrence \cite{frank2004data}. A vast number of these algorithms are used in the articles surveyed including different types of Neural Networks (NN), as seen in Table~\ref{tab:tools}; however, the most common is Support Vector Machines (SVM). SVM are frequently used in biomedical applications, and this survey shows that the addition of GA does not change this fact. One of the biggest appeal of SVM is their near perfect success rate and their perceived simplicity of simply assigning labels to objects based on what side of a hyperplane they end up on \cite{noble2006support}. Computation requirements for the SVM scale quadratically, resulting in longer run times as data inputs increase \cite{noble2006support}. This in itself is not necessarily a current negative; however, as applications become more complex, the SVM quadratic run time growth should not be ignored in future works employing it alongside GA.    

\section*{3 Performance Metrics}
A key step in the process of building a machine learning model is to estimate its performance on data that was not part of building the model. The data to evaluate the performance of the model is called the testing set, while the data that is used to build the model is called the training set. A primary concern for any machine learning prediction model is avoiding a model with either high bias or high variance. Bias is the error resulting from a wrong assumption. A model with high bias oversimplifies. This is also known as underfitting. It results in a large error between the test set outcome value and the model prediction. Variance is the error from the model being overly sensitive to fluctuations in the training set. High variance can cause an algorithm to model the noise in the data, which results in model overfitting. High variance decreases the amount of flexibiliy, and reduces the ability of the model to generalize to unseen instances. A visualization of the trade-offs made between bias and variance can be seen in Figure~\ref{fig:under_over_fit}. 

The confusion matrix is a key concept related to the performance metrics of a classifier model. The confusion matrix is simply a square matrix that records the counts of the true positive (TP), true negative (TN), false positive (FP), and false negative (FN) predictions of a classifier. The true positive rate (TPR) is calculated as the number of true positives divided by the sum of the false positives and the true negatives, \begin{equation} \hspace{3cm} TPR ={\frac{\displaystyle TP}{\displaystyle FN + TP}} \end{equation} The false positive rate (FPR) is calculated as the number of false positives divided by the sum of the false positives and the true positives, \begin{equation} \hspace{3cm} FPR ={\frac{\displaystyle FP}{\displaystyle FP + TN}} \end{equation} A dimension of the confusion matrix represents the instances in a predicted class while the other dimension represents the instances in the actual class (ground truth). If the predicted class is the same as the ground truth, then the confusion matrix will label this sample as true, otherwise false \cite{chawla2009data}. 

The precision is defined as the ratio of the true positives to the sum of the true positives and the false positives,  \begin{equation} \hspace{3cm} Precision ={\frac{\displaystyle TP}{\displaystyle TP+FP}} \end{equation} The recall is defined as the ratio of the true positives to the sum of the true positives and the false negatives, \begin{equation} \hspace{3cm}Recall ={\frac{\displaystyle TP}{\displaystyle TP+FN}} \end{equation}The F\textsubscript{1} score is defined as the two divided by the inverse of the precision, plus the inverse of the recall, \begin{equation} \hspace{3cm}F\textsubscript{1} ={\frac{\displaystyle 2}{\displaystyle recall\textsuperscript{-1} + precision\textsuperscript{-1}}} \end{equation}
	
Receiver Operating Characteristic (ROC) graphs are useful tools to select models for classification based on performance with respect to the false positive rate (FPR) and true positive rate (TPR), which are computed by shifting the decision threshold of the classifier. The diagonal of an ROC graph presents random guessing (50 percent probability of being correct), and classification models that fall below this value are considered worse than random guessing. A perfect classifier would fall into the top left corner of the graph with a TPR of 1 and an FPR of 0. Based on the ROC curve, the area under the curve can be computed to characterize the performance of the classification model \cite{chawla2009data}. 

The prediction error and accuracy provide general information regarding the performance of the prediction model. The error can be understood as the sum of the false predictions divided by the total number of predictions, \begin{equation}  \hspace{3cm}Error ={\frac{\displaystyle FP+FN}{\displaystyle FP+FN+TP+TN}} \end{equation} The accuracy is calculated as the sum of the correct predictions divided by the total number of predictions. More precisely, accuracy is the ratio of the number of correct predictions (the sum of the true positives and true negatives) to the total number of predictions from the model (the sum of the true positives, true negatives, false positives, false negatives), \begin{equation}  \hspace{3cm}Accuracy ={\frac{\displaystyle TP+TN}{\displaystyle FP+FN+TP+TN}} \end{equation}
	
There are many methods to evaluate the performance of a model. Each performance metric has certain advantages and disadvantages based on the data, such as the number of classes in the prediction variable, the number of instances of each class, or how imbalanced the outcome class happens to be, and the cost of misclassifying a prediction. In medicine, misclassification can be deadly. The discussion relating to advantages and disadvantages will focus on the accuracy, as it was the most common performance metric. Some attention will be also be paid to the true positive rate and false positive rate, as it offers a more nuanced metric, especially in relation to biomedical applications. What metrics are used by each surveyed paper can be found in Table \ref{tab:metrics}.

\subsection*{Advantages}

Accuracy is a simple performance metric to compute, and the most intuitive evaluation method. It is the most common metric, so it is often used to compare with other models in the literature. 
	
The true positive rate and false positive rate are especially usefully for imbalanced class problems. For example, in tumour diagnosis, the detection of malignant tumours is the primary concern since missing the potential presence of a tumour could have serious implications, like death. However, it is also important to decrease the number of benign tumours that are incorrectly classified as malignant (false positive) to not unnecessarily concern a patient. The true positive rate provides useful information about the fraction of positive (or relevant) samples that were correctly identified out of the total number of positives. In medicine, the samples tend to be imbalanced, so the true positive rate and false positive rate will be the most appropriate performance metric. 
	
An ROC graph is a useful tool to visualize the true positive rate and false positive rate. Finding the area under the curve is a simple method to determine the performance of the model.

\subsection*{Disadvantages}

Accuracy was the primary performance metric used in this scoping review. However, it has some limitations that are important to consider, especially in the medical domain.  It is only a reliable performance metric when the number of samples are equal for each class (no imbalance). For example, consider a case where 99 percent of samples belong to class A and only 1 percent to class B. Then it is trivial for the model to obtain 99 percent  accuracy by simply predicting every training instance to belong to class A. If the identical model is evaluated on a different test set then the accuracy would be significantly reduced. For example, if the test set has 60 percent of its samples from class A and 40 percent of its samples from class B, then the accuracy would plummet to 60 percent. This examples illuminates the potential for the accuracy metric to be misleading, which can lead to assuming the model is better than reality. In the medical field, the price of misclassifying a sample has the potential to be extremely costly. If the model is attempting to predict a rare but fatal disease, the cost of failing to diagnose the disease of a sick person is much greater than the cost of sending a healthy person to do more tests.

The papers mostly failed to evaluate a major drawback of GA, which is the amount of computation it requires. In traditional machine learning, such as neural networks, the model improves as the amount of training data increases. However, the performance of a GA might degrade before it improves. GAs also keep a population of solutions, instead of a single solution. These requirements of GA are computationally costly, and should be evaluated as a performance metric whenever considering a genetic algorithm as a learning algorithm\cite{mitchell1998introduction}.

\section*{4 Discussion and Future Research Directions}

Some of the founders of computer science, such as Alan Turing, John von Neumann, Norbert Wiener, were motivated by the idea of providing computer programs with operations like self-replication and adaption\cite{mitchell1998introduction}. These motivations have been explored in various areas of research such as evolution strategies, evolutionary programming, and genetic algorithms. These efforts grew into the field known as evolutionary computation, of which GAs are the most prominent example. 

The GAs are a powerful tool for solving problems and for simulating natural systems in a wide range of scientific fields. GAs are promising approaches for solving challenging technological problems. GAs are an important area of research in machine learning, especially working together with other approaches such as neural networks. GAs are part of a movement in computer science that explores biologically-inspired approaches to computation. These systems are adaptable, parallel, able to handle complexity, able to learn, and even be creative \cite{mitchell1998introduction}. Furthermore, the computing resources that are currently widely available and allow for unprecedented parallel processing are well-suited to implementing GA. 

 The GA attempt to model natural evolution, which is done with operators such as adaption, selection, crossover, and mutation. This approach retains a population of solutions that converges on the objective, which is a form of black-box optimization. However, natural evolution is a  process that ceaselessly creates greater complexity and novelty, rather than a process that converges on a single solution. In fact, evolution on Earth can be thought of as a single run of a single algorithm that invented all of nature \cite{lehman2011abandoning}. Another term for the notion of a single process inventing massive complexity for near-eternity is “open-ended.” Open-endedness has proven impossible to program. Presently, no such algorithm exists that has the endless, prolific creative potential of natural evolution.

Currently, most evolutionary algorithms (EAs) converge to a solution, based on the fitness function that is chosen. The fitness function, which tends to select the “best" performing individuals in the population of solutions, acts as an objective that is optimized. The optimization consists of selecting more of the fitter solutions on average, while only selecting a minority of other “less" fit solutions to maintain some diversity. However, the divergence of natural evolution and the “open-endedness” is not implemented with this approach. Natural evolution is not structured like an optimization algorithm as there is no explicit objective, and organisms are often rewarded for being different rather than just better. For example, organisms that are sufficiently different from their predecessors can establish a new niche in which they can benefit from reduced competition and are therefore more likely to survive \cite{kirschner1998evolvability}\cite{lehman2013evolvability}.  In opposition to optimization algorithms that converge to a single “best” solution, natural evolution has a tendency toward divergence. This alternative perspective in evolutionary computation in that evolution is an algorithm for diversification rather than optimization \cite{pugh2016quality}.

An EA inspired by this approach is called novelty search (NS), which searches for behavioural diversity without any explicit objective. In some domains, NS finds the global optimum even when objective-based solutions consistently fail \cite{pugh2016quality}. An algorithm that avoids an objective function is able to find solutions that are not possible if attempting to solve them directly with objectives. This insight has implications beyond GA, such as in the pursuit of ``human-level" AI, since it captures what many consider our most human-like quality--creativity.

A potentially fruitful application for open-ended evolutionary algorithms is in any sort of creative design. This includes the design of cars, art, medicines, robots, video games, and so on. Open-ended evolutionary algorithms offer the potential to generate endless alternatives in almost any conceivable design domain, in the same way that natural evolution generated endless solutions to the problems of survival and reproduction in nature \cite{lehman2011abandoning}.

There are many potential biomedical applications for open-ended, evolutionary algorithms.  One would be the development of vaccines. The open-ended algorithm could search the space of possibilities while simultaneously finding solutions that work in each environment. Provided some initial set of rules that describe what is possible biologically, the algorithm could continuously explore this space of possibilities, and report any number of potentially useful findings to researchers to  investigate further.   
   
\section*{5 Conclusion}

Population-based search like GAs are often combined with other machine learning algorithms. In classification problems, GA serves as a population of solutions, rather than a single solution. In this scoping review, GAs combined with Support Vector Machines were found to perform best. The performance metric that was evaluated most often was the accuracy. This avoids measuring the main weakness of GA, which is computational time. In an attempt to better utilize the power of GAs, the future of GAs could be “open-ended” evolutionary algorithms, which attempt to increase complexity and find diverse solutions, rather than optimize a fitness function to find a single “best” solution. This approach attempts to model the most powerful feature of natural evolution—its endless ability to create novel and creative solutions to fit an environment that is constantly changing.


\begin{backmatter}

\section*{Competing interests}
  The authors declare that they have no competing interests.

\section*{Author's contributions}
The first three authors contributed equally for the development of this research article. The last author provided supervision and guidance.

\section*{Acknowledgements}
The authors would like to thank the infrastructure support provided by the CASES Building at Lakehead University. 

\bibliographystyle{bmc-mathphys} 
\bibliography{bmc_article}      





\begin{figure}[h!]
    \centering
    \includegraphics[width=0.5\textwidth]{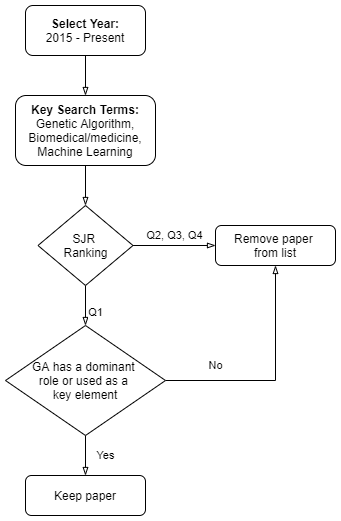} 
    \caption{\csentence{Finding Quality Papers.}
      Search criteria used to identify papers for the article.}    \label{fig:search_diagram}
\end{figure}

\begin{figure}[h!]
    \centering
    \includegraphics[width=0.9\textwidth]{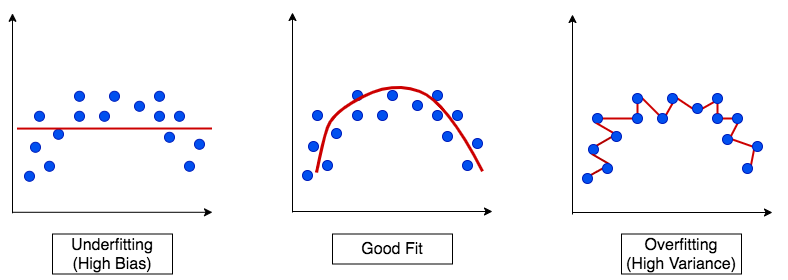} 
    \caption{\csentence{Bias Variance Trade-Offs}
      Visualization of underfitting and overfitting. }    \label{fig:under_over_fit}
\end{figure}



\begin{center}
\begin{longtable}{| p{0.85cm} | p{0.5cm} | p{0.5cm} | p{0.9cm} | p{3cm} | p{3.5cm} | p{2.5cm} |}
\caption{Information about the Quality of the Papers\label{tab:papers}}\\
\hline
 Article & SJR Rank & Cities Per Year & Year of publication & How the GA is Used & Benefits & Biomedical Applications\\
 \hline
 \cite{salem2017classification} & Q1 & 23 & 2017 & Genetic Programming used for cancer disease classification. & IG/GA method improves classification accuracy by reducing the number of features and preventing the GA from being trapped by local optimum. & Cancer Classification\\
\hline
\cite{al2017examining} & Q1 & 6.67 & 2017 & Using GA for feature selection. 
Combing GA and PSO for feature selection. 
Classification using GP. &
Selecting fewer genes, classification algorithm takes less computational time. 
GA/DT and GA/GP yields highest classification accuracy.
 & Colon Cancer. \\
\hline
\cite{lu2017hybrid} & Q1 & 28.33 & 2017 & Adaptive Genetic Algorithm (AGA) improves conventional GA by adjusting values of crossover and mutation probability.  The adaptability increases robustness, increasing the chance of finding optimal solutions. 
& 
Combing MIM (Mutual information maximization) with AGA, eliminates redundant samples and reduces the dimension of the gene expression data. & General applications to biomedical datasets. \\
\hline
\cite{alivckovic2017breast}& Q1 & 27.67 & 2017 & GA feature selection - extraction of information and significant features.  & Reduces computation complexity and speeds up the data mining process.

GA for feature selection, combined with Rotation Forest resulted in highest classification accuracy. 
& Breast Cancer Diagnosis. \\
\hline
\cite{gangavarapu2019novel} & Q1 & 2.00 & 2019 & GA optimizes the subspace ensembling process. 
&
Optimizing with GA, outperforms selected base feature selection techniques in terms of prediction accuracy. &
General applications to biomedical datasets.\\
\hline
\cite{mohammed2018real} & Q1 & 18.50 & 2018 & Machine learning approaches based on the GA for feature selection. & Reduces overlapping between classes, and reduces the number of features to enhance the time cost. & Visualizing border points for resection of Nasopharyngeal Carcinoma. \\
\hline
\cite{khanduzi2019fast} & Q1 & 2.00 & 2019 & GA results in high-quality solutions (accuracy and execution time) & GA-FBC (Fast Branch Cut Method) provides efficient solutions, regarding performance metrics.  & Biomedical supply chain networks.\\
\hline
\cite{subasi2019epileptic}& Q1 & 37.00 & 2019 & GA used to determine optimum parameters of SVM. & Combing GA with SVM offers quick global optimizing ability.  & Classification of EEG data for Epileptic seizure detection.\\
\hline
\cite{soufan2015dwfs} & Q1 & 13 & 2015 & Feature selection tool developed based on GA & Able to significantly reduce the number of features without sacrificing classification performance. & 
Feature selection for biomedical data.\\
\hline
\cite{li2016genetic} & Q1 & 9.25 & 2016 & Uses a GA-based weighted ensemble method to predict transposon-derived piRNAs & Has higher performance and robustness compared to similar methods. & Prediction of piRNAs.\\
\hline
\cite{tangherloni2019genhap}  & Q1 & 8.00 & 2019 & GAs with tournament selection and elitism  & Speeds up the required computations, and can take into account datasets produced by 3rd generation sequencing technologies & Helps solve the haplotyping problem.\\
\hline
\cite{moraes2017gass} & Q1 & 1.33 & 2017 & GA performs the search of the generated database & A freely available method, through a webapp that ranks among the top (4th) & Identification of enzyme active sites allowing for non-exact matches.\\
\hline
\cite{zhang2019selection} & Q1 & 1.00 & 2019 & GA used to find subset of the principal components, from a Principal component analysis. & Use of Principal component analysis before the GA improves the results of GA selection & Help identify what treatments should be done for different patients.\\
\hline
\cite{ramadan2016protein} & Q1 & 3.50 & 2016 & GA used to identify complexes in protein interaction networks & Method allows for identifying clustering with varying densities. It is more scalable and robust and it can be tuned. & Used to detect dense and sparse protein clusters.\\
\hline
\cite{sayed2019nested} & Q1 & 11.00 & 2019 & Uses 2 GAs. The outer GA serves as the main algorithm and outputs the subset of genes evaluated by SVM. The inner GA takes data from DNA methylation and outputs subset of CpG sites. & Far higher accuracy compared to other methods, and has been shown to be able to differentiate between lung cancer subtypes & Identification of disease (cancer)  biomarkers.\\
 \hline
 \cite{lee2018comprehensive} & Q1 & 4.00 & 2018 & Compares the performance of multiple GAs & N/A & Guidelines for the development of GA based solutions for DNA motif prediction.\\
 \hline
 \cite{oztekin2018decision} & Q1 & 17 & 2018 & GA used in feature selection while predicting the Quality of Life & Study included all UNOS features (after preprocessing) allowing for their effect to be assessed.  & Minimize or eliminate personal bias in lung transplants by automation. Helping to increase the rate of successful lung  transplants.  \\
\hline
\cite{corus2017standard} & Q1 & 10.67 & 2018 & Prove the benefits of crossover in Genetic Algorithms & Established that GA with crossover is 25 percent faster than mutation alone, with certain parameters. & N/A\\
\hline
\cite{hashem2017comparison} & Q1 & 2.67 & 2018 & Finding the best features, predict advanced fibrosis. & GA is able to work in parallel. & Predict advanced fibrosis.\\
\hline
\cite{ansotegui2015model} & Q1 & 10.20 & 2015 & Automatic algorithm configuration & Numerical results show that model-based genetic algorithms significantly improve our ability to effectively configure algorithms automatically. & N/A\\
\hline
\cite{ghosh2016incorporating} & Q1 & 10.00 & 2016 & GA for combining representations of learned information such as known shapes, regional properties and relative position of objects into a single framework to perform automated three-dimensional segmentation. & GA-based method are very useful for medical imaging applications. & GA tested for prostate segmentation on pelvic computed tomography and magnetic resonance images.\\
\hline
\cite{hemanth2019modified} & Q1 & 5.00 & 2018 & Three different modified Genetic Algorithm approaches are proposed  for feature selection & The number of features are reduced, decreasing the dimensionality of the features & Magnetic Resonance brain image classification \\
\hline
\cite{bhardwaj2016novel} & Q1 & 8.25 & 2016 & Classification & Proposes a constructive genetic programming approach that increasing the number of useful “building blocks” & Classifying EEG signals\\
\hline
\cite{tan2016genetic} & Q1 & 1.50 & 2016 & Feature Selection & Compared the performance to support vector machines, logistic regression and performed better. & Recognition of cancerous cells and also gene expression profiling data\\
\hline
\cite{tan2018genetic} & Q1 & 1.00 & 2018 & Proposes a hybrid of a generic algorithm and fuzzy logic for pattern recognition & Exemplifies the advantage of the best heuristic search (genetic algorithm) with the ease of understanding and interpretability of fuzzy logic & Breast Cancer, Diabetes, Parkinson’s Disease\\
\hline
\cite{dashtban2017gene}& Q1 & 17.00 & 2017 & Introduces a novel evolutionary algorithm, Intelligent Dynamic Genetic Algorithm (IDGA), based on the GAs and Artificial Intelligence. & Reduces dimensionality of the feature space to provide statistically important features.  & Gene Selection for Cancer Classification \\
\hline

\cite{liu2019deep}& Q1 & 3.00 & 2018 & Optimized GA-based strategy to explore CNN structures. &  GA based network evolution approach to search for the fittest genes to optimize network structure automatically. Outperforms  state-of-the-art methods consistently at various noise levels. & Medical Image De-Noising \\
\hline
\cite{fogue2016non} & Q1 & 4.5 & 2016 & GAs assign services/tasks to ambulances with the aid of local search and the constraint-dominance concept   & Program is shown to reduce waiting time by 10\% and increase vehicle usage by 30\% & Route planning for ambulances responding to non-emergency assignments \\
\hline
\cite{plawiak2018novel} & Q1 & 38 & 2018 & GA used for feature selection and classifier parameter optimization &  Focuses on efficiency while retaining accuracy & Uses ECG signal to efficiently classify cardiac disorders \\
\hline
\cite{la2019multidimensional} & Q1 & 15 & 2018 & Multiclass classification method that learns multidimensional feature transformation using Genetic Programming & Optimizes models by first performing a transformation of the feature space into a new space of potentially different dimensionality, and then performing classification using a distance function in the transformed space & Identifying nonlinear interactions in simulated genome-wide associated studies\\
\hline
\cite{devarriya2020unbalanced} & Q1 & 2 & 2019 & Classification  & The GA is better able to handle the unbalanced dataset by altering the fitness function 
& Breast Cancer classification \\
\hline
\end{longtable}
\end{center}

\begin{center}
\begin{longtable}{| p{1.5 cm} | p{6cm} | p{5.5cm} |}
 \caption{Datasets Used\label{tab:data}}\\
\hline
 Article & Dataset Used & Source\\
\hline
\cite{salem2017classification} & 7 Skewed Gene Expression Datasets: 

Leukemia, Colon tumor, Central nervous system, Lung cancer-Ontario, Lung cancer-Michigan, Diffuse Large B-Cell Lymphoma (DLBCL) and Prostate Cancer & Kent Ridge Biomedical Dataset website\\
\hline
\cite{al2017examining} & Gene expression dataset - Colon Cancer & Extracted from public cancer datasets \\
\hline
\cite {lu2017hybrid} & 6 Gene Expression Datasets: 

Leukemia, Colon, Prostate, Lung, Breast, SBRCT (Small Blue Round Cell Tumor) & Does not specify.\\
\hline
\cite{alivckovic2017breast} & 2 different Wisconsin Breast Cancer datasets:

Wisconsin Breast Cancer (Diagnostic) (WBC (DIAGNOSTIC), Wisconsin Breast Cancer Original Dataset & Obtained from UCI Machine Learning Repository 

https://archive.ics.uci.edu/ml/index.php\\
\hline
\cite{gangavarapu2019novel} & 3 Biomedical Datasets: 

1. Translation Initiation Sites (TIS) 

2. Skin Cancer

3. Epileptic Seizure Recognition & 1. Extracted from genome sequences from the GenBank

https://www.ncbi.nlm.nih.gov/genbank/

2. 
Extracted from pixel information of 28 x 28RGB images of skin cancer MNIST: HAM10000 

3. EEG Segments\\
\hline
\cite{mohammed2018real} & Dataset included 381 NPC endoscopic images with 159 tumors (abnormal cases) and 222 of normal tissues & 
NPC endoscopic images obtained from ENT Department Tun Fatimah Specialist Hospital, Muar, Johor\\
\hline
\cite{khanduzi2019fast} & N/A & N/A\\
\hline
\cite{subasi2019epileptic} & Five EEG Datasets & http://www.meb.unibonn.de/epileptologie/  science/physik/eegdata.html\\
\hline
\cite{soufan2015dwfs} & 9 datasets in experimental setup:

TIS, TFTF, Medelon, Wdbc, Pre-miRNAs, Lung cancer (microarrays), Leukemia (microarrays), Prostate cancer (microarrays), Promoters & http://www.cbrc.kaust.edu.sa/dwfs/data\_desc.php \\
\hline
\cite{li2016genetic} & 6 constructed datasets half balanced, half imbalanced for: Human, Mouse and Fruit Fly data. & -Constructed Datasets at: https://github.com/zw9977129/piRNAPredictor

-NONCODE V3 at: https://www.ncbi.nlm.nih.gov/pmc/articles/  PMC3245065/  

-UCSC Genome Browser at: https://www.ncbi.nlm.nih.gov/pubmed/24270787

-NCBI Gene Expression Omnibus at: https://www.ncbi.nlm.nih.gov/pubmed/15608262 and https://www.ncbi.nlm.nih.gov/pubmed/  17952056/\\

\hline
\cite{tangherloni2019genhap} & 2 generated using the: reference sequence of the human chromosome 22 & -UCSC Genome Browser at: GRCh37/hg19Feb.2009assembly 

-Real Dataset from: https://www.pacb.com/blog/data-release-54x-long-read-coverage-for/ 

-Tested models found at: https://github.com/andrea-tango/GenHap/blob/master/Models.zip\\
\hline
\cite{moraes2017gass} & Constructed database consisting of data from the Protein Data Bank & Protein Data Bank: https://www.rcsb.org/ 

-Templates from: Catalytic Site Atlas found at: https://www.ncbi.nlm.nih.gov/pubmed/24319146  

-Enzymes from the NCBI-VAST non-redundant database found at: https://www.ncbi.nlm.nih.gov/pubmed/24319143\\
\hline
\cite{zhang2019selection} & Constructed dataset using collected samples, Greengenes 16S taxonomy database V13.5 &  Can be requested from Menzies Health Institute Queensland.\\
\hline 
\cite{ramadan2016protein} & Collins protein interaction network from the  the BioGrid dataset, MIPSyeast genome database, CYC2008 & http://faculty.kfupm.edu.sa/ics/eramadan/  GACluster.zip \\
\hline 
\cite{sayed2019nested} & TCGA gene expression data: DNA Methylation, GEO gene expression data, Copy Number Variation  (CNV) & -The Cancer Genome Atlas found at: https://tcga-data.nci.nih.gov/tcga, 

-GEO from NCBI found at: https://www.ncbi.nlm.nih.gov/gds 

-CNV from FireBrowse found at: http://firebrowse.org/?cohort=COAD\\
\hline
\cite{lee2018comprehensive} & Used to compare GAs: CRP(18), CREB(17), SRF(20), ERE(25), MEF2(17), MYOD(17), TBP(39), E2F(25). Numerous others mentioned for each individual GA & N/A\\
\hline
\cite{oztekin2018decision} & Dataset constructed from UNOS standard Transplant Analysis and Research files for: lung & Must be requested from the: United Network for Organ Sharing\\
\hline
\cite{corus2017standard} & N/A & N/A \\
\hline
\cite{hashem2017comparison} & A group of 39,567 chronic hepatitis C patients from the National Treatment Program of HCV patients in Egypt  & Egyptian National Committee for Control of Viral Hepatitis database \\
\hline
\cite{ansotegui2015model} & N/A & N/A \\
\hline
\cite{ghosh2016incorporating} & Pelvic images were obtained from CT and MRI scans of patients being treated for Prostate Cancer at Oregon Health and Science University,  CT and MRI images of 10 patients manually segmented by Dr. A. H. and Dr. J. T, (Dept. of Radiation Medicine, OHSU) & Oregon Health and Science University (OHSU) \\
\hline
\cite{hemanth2019modified} & Real time abnormal brain tumour images are used in this work. These images are collected from M/s. Devaki Scan centre, Madurai, India & N/A \\
\hline
\cite{bhardwaj2016novel} & EEG Dataset & N/A\\
\hline
\cite{tan2016genetic} & A total of 31 oral cancer cases of 3-year prognosis & The Malaysia Oral Cancer Database and Tissue Bank System (MOCDTBS) coordinated by the OCRCC, Faculty of Dentistry, University of Malaya \\
\hline

\cite{tan2018genetic} & 2 Datasets:
Wisconsin Breast Cancer (458 Benign, 241 Breast Cancer), Pima Indian Diabetes (500 without diabetes, 268 with diabetes) 
& https://archive.ics.uci.edu/ml/datasets/
 Breast+Cancer+Wisconsin+(Diagnostic)
\\

\hline
\cite{dashtban2017gene} & 1. Small Round Blue Cell Tumor

2. Breast Cancer

3. Large B-cell lymphoma - Standford University

4. Acute Lymphoblastic Leukemia / Acute Myeliod Leukemia 

5.Prostate Cancer 
 & 
 1,2 and 5. does not specify. 
 
 3. Standford University 
 
 4. Broad Institute Website
 
 https://www.broadinstitute.org/\\
\hline
\cite{liu2019deep} & Collection of 10,775 cerebral perfusion CT images 
&  Created own data sets.\\
\hline
\cite{fogue2016non} & 3 scenarios using provided data from actual events & Ambulance Company located in Huesca, Spain \\
\hline
\cite{plawiak2018novel} & Constructed database using ECG signals from 45 patients using data from the MIH-BIH Arrhythmia database & MIH-BIH Arrhythmia database accessed through the PhysioNet service. Constructed database provided at: https://www.sciencedirect.com/science/
article/abs/pii/S0957417417306292?via\%3Dihub \\
\hline
\cite{la2019multidimensional} & 16 GAMETES datasets & https://www.ncbi.nlm.nih.gov/pubmed/23025260 \\
\hline
\cite{devarriya2020unbalanced} & Wisconsin Breast Cancer dataset & https://archive.ics.uci.edu/ml/datasets/Breast+ Cancer+Wisconsin+(Diagnostic)  \\
\hline
\end{longtable}
\end{center}

\begin{center}
\begin{longtable}{| p{1.5 cm} | p{1.5cm}| p{3cm} |}
 \caption{Analysis for Reproducibility\label{tab:reproduce}}\\
\hline
 Article & Pseudocode & Public Code Repository \\
\hline
\cite{salem2017classification} & $\checkmark$ & $\times$ \\
\hline
\cite{al2017examining} &$\times$ &$\times$ \\
\hline
\cite{lu2017hybrid} &$\times$ &$\times$ \\
\hline
\cite{alivckovic2017breast} &$\times$ &$\times$ \\
\hline
\cite{gangavarapu2019novel} &$\checkmark$ &$\times$ \\
\hline
\cite{mohammed2018real} &$\times$ &$\times$ \\
\hline
\cite{khanduzi2019fast} &$\checkmark$ &$\times$ \\
\hline
\cite{subasi2019epileptic} &$\times$ &$\times$ \\
\hline
\cite{soufan2015dwfs} &$\times$ &$\times$ \\
\hline
\cite{li2016genetic} &$\times$ &$\checkmark$ \\
\hline
\cite{tangherloni2019genhap} &$\times$ &$\checkmark$ \\
\hline
\cite{moraes2017gass} &$\times$ &$\times$ \\
\hline
\cite{zhang2019selection}&$\times$ &$\times$ \\
\hline
\cite{ramadan2016protein} &$\checkmark$ &$\checkmark$ \\
\hline
\cite{sayed2019nested} &$\checkmark$ &$\times$ \\
\hline
\cite{lee2018comprehensive} &$\times$ &$\times$ \\
\hline
\cite{oztekin2018decision} &$\times$ &$\times$ \\
\hline
\cite{corus2017standard} &$\checkmark$ &$\times$ \\
\hline
\cite{hashem2017comparison} &$\times$ &$\times$ \\
\hline  
\cite{ansotegui2015model} &$\times$ &$\times$ \\
\hline
\cite{ghosh2016incorporating} &$\times$ &$\times$ \\
\hline
\cite{hemanth2019modified} &$\times$ &$\times$ \\
\hline
\cite{bhardwaj2016novel} &$\checkmark$ &$\times$ \\
\hline
\cite{tan2016genetic} &$\times$ &$\times$ \\
\hline
\cite{tan2018genetic} &$\checkmark$ &$\times$ \\
\hline
\cite{dashtban2017gene} &$\times$ &$\times$  \\
\hline
\cite{liu2019deep} & $\checkmark$ &$\times$  \\
\hline
\cite{fogue2016non} &$\times$ &$\times$  \\
\hline
\cite{plawiak2018novel} &$\checkmark$ &$\times$  \\
\hline
\cite{la2019multidimensional} &$\times$ & $\times$ \\
\hline
\cite{devarriya2020unbalanced} & $\checkmark$& $\times$ \\
\hline
\end{longtable}
\end{center}

\begin{center}
\begin{longtable}{| p{1.5 cm} | p{4.5cm}| p{6cm} |}
 \caption{Tools Used\label{tab:tools}}\\
\hline
 Article & Tools & Additional ML Algorithms Utilized/Validation\\
\hline
\cite{salem2017classification} & Does not specify. & 10-fold cross validation

Classification Algorithm: 

- Genetic Programming (GP)\\
\hline
\cite{al2017examining} & Weka Machine Learning package & Leave one out cross validation (LOOCV), k-fold cross validation

Classification Algorithms: 

- Decision Tree,

- Naive Bayes, 

- Support Vector Machine, 

- Genetic Programming\\
\hline
\cite{lu2017hybrid} & Does not specify. & 
Multiple cross validations. 

Classification Algorithm:

- Back Propagation Neural Network (BP), 

- Support Vector Machine (SVM), 

- Extreme Leaning Machine (ELM), 

- Regularized Extreme Leaning Machine (RELM)\\
\hline
\cite{alivckovic2017breast} & Weka employed to implement algorithms. & 

10-fold cross validation.

Classification Algorithm: 

- Rotation Forest Model, 

- Logistic Regression, 

- Bayesian Network, 

- Multilayer Perceptron (MLP), 

- Radial Basis Function Networks (RBFN), 

- Support Vector Machine (SVM), 

- C4.5 Decision Tree, 

- Random Forest, 

- Rotation Forest\\
\hline
\cite{gangavarapu2019novel} & All experiments coded in Python 2.7 and Weka 3.8.3 (to implement all the predetermined feature selection methods). 

Python Scikit-learn package implemented all the classifiers. & 10-fold cross validation

Classification Algorithms: 

- Random Forests, 

- Bootstrap Aggregating with C4.5 Decision Trees, 

- K-Nearest Neighbour\\
\hline
\cite{mohammed2018real} & MATLAB 2014a utilized for the evaluation of the present approach. & Cross validation.

Classification Algorithms:

- Artificial Neural Networks \\
\hline
\cite{khanduzi2019fast} & All approaches in this study are coded using MATLAB software.  & N/A\\
\hline
\cite{subasi2019epileptic} & Does not specify. & 

10-fold cross validation
s
Classification Algorithm:

- Support Vector Machine \\
\hline
\cite{soufan2015dwfs} & PGAPack software libraries, K-Nearest Neighbour from AlgLib Library, Matlab R2012b & Classification algorithms :

- K-Nearest Neighbour

- Naive-Bayes

- Combination of above 2 algorithms.\\
\hline
\cite{li2016genetic} & Random forest classification engine from scikit-learn python package & 10-fold cross validation, 
Their weighted ensemble method is constructed using training data.

Classification Algorithms:
- Random forest

- Support Vector Machine \\
\hline
\cite{tangherloni2019genhap} & Message Passing Interface specifications in C++, Roche/454 genome sequencer, PacBio RS II sequencer, General Error-Model based SIMulator toolbox  & N/A \\
\hline
\cite{moraes2017gass} & Flask framework for Python, frontend developed using Bootstrap framework. Runs on top of an Apache server with communication being made using a Web Server Gateway Interface & N/A \\
\hline
\cite{zhang2019selection} & 
Use of sequence analysis pipelines such as: 

- DADA2

- PEAR Software V0.9.6

- BWA Software Package V0.7.12

- Stats package in R &
5-fold Cross Validation

- Classification Algorithms:

- Logistic Regression\\
\hline
\cite{ramadan2016protein} & GO term finder & Spectral clustering\\
\hline
\cite{sayed2019nested} & - biomaRt 

- GenomicRanges

- Mminfi

- IlluminaHumanMethylation27kabbi:ilmn12:hg19 R packages

- SVM method from e1071 package

- Gene Ontology, Kyoto Encyclopedia of Genes and Genomes &
5-fold Cross Validation

Deep-learning Neural Network

Classification algorithm:

- Support vector machine\\
\hline
\cite{lee2018comprehensive} & Local Search Techniques 

Gibbs Sampling

Expectation maximization 

Additional non-GA methods/tools mentioned but not shown to be tested: list can be found in supplementary materials pdf. & 
GA Motif discovery:
PCEA, GAPWM, kmerGA, GAMI, FGMA, Paul and Iba, Gadem, GA-DPAF, GASMEN, MDGA, GALF (GALF-P), GALF-G, GAME, GEMFA, GAPK, iGAPK\\
\hline
\cite{oztekin2018decision} & Does not specify. & 5-fold Cross Validation

Random undersampling

Classification algorithm used:

- k-Nearest neighbour

- Support Vector Machine (SVM)

- Artificial Neural Network (ANN)\\
\hline
\cite{corus2017standard} & The ONEMAX benchmark function & N/A.\\
\hline
\cite{hashem2017comparison} & MedCalc, MATLAB, Weka & Implemented several types of Machine learning techniques: 

- particle swarm optimization

- multi-linear regression

- decision tree learning algorithms to compare.\\
\hline
 \cite{ansotegui2015model} & Comparing Continuous Optimizers (COCO) software & 
 Classification Algorithm:
 
 - Random Trees \\
 \hline
 \cite{ghosh2016incorporating} & In preprocessing, the images were improved with the “imadjust” function in MATLAB & N/A \\
 \hline
 \cite{hemanth2019modified} & Implemented in MATLAB & Neural Network\\
 \hline
 \cite{bhardwaj2016novel} & N/A & N/A\\
 \hline
 \cite{tan2016genetic} & GPLAB, which  is a genetic program- ming toolbox, which runs in the MATLAB environment & 
 Classification Algorithms:
 
 - Support Vector Machine
 
 - Logistic Regression  \\
 \hline
 \cite{tan2018genetic} & N/A & Fuzzy Logic\\
 \hline
 \cite{dashtban2017gene} & Does not specify & LOOCV and 10 Fold CV

Classifiers:

- KNN

- Support Vector Machine

- Naive Bayes

Filter Methods: 

- Laplacian-score

- Fisher-score 
 \\
 \hline
 \cite{liu2019deep} & GA progress is processed on Tensorflow platform with GEFORCE GTX TITAN GPUs  &  Convolutional Neural Networks
\\
 \hline
 \cite{fogue2016non} & Google Maps API & N/A \\
 \hline
 \cite{plawiak2018novel} & MATLAB R2014b, libsvm library for MATLAB &  4-fold cross validation
 
 10-fold cross validation
 
 Classification Algorithms:
 
 -Support Vector Machine
 
 -K-Nearest Neighbour
 
 -Probabilistic Neural Network 
 
 -Radial Basis Function Neural Network\\
 \hline
 \cite{la2019multidimensional} & PyTorch & Neural Network, Decision Tree \\
 \hline
 \cite{devarriya2020unbalanced} & Python packages &  None\\
 \hline
\end{longtable}
\end{center}

\begin{center}
\begin{longtable}{| p{0.7cm}| p{0.5cm}| p{0.5cm}| p{0.5cm}| p{0.5cm}| p{0.5cm}| p{0.5cm}| p{0.5cm}| p{1cm}| p{1cm}| p{0.5cm}| p{0.5cm}| p{0.5cm}| p{0.5cm}| p{1.5cm}| p{1cm}|}
 \caption{Performance Evaluation\label{tab:metrics}}\\
\hline
 Article & Acc. & ROC Curve & AUC & TP & TN & FP & FN & Specificity & Sensitivity / Recall & Prec./ PPV & F- Measure & Avg. Runtime & Comp. Complexity & Other  \\
\hline
\cite{salem2017classification} & $\checkmark$ & $\times$ & $\times$ & $\checkmark$ &$\checkmark$ &$\checkmark$ &$\checkmark$ &$\checkmark$ &$\checkmark$ &$\times$ &$\times$ &$\times$ &$\checkmark$ & \\
\hline
\cite{al2017examining} & $\checkmark$ &$\times$&$\times$&$\times$ &$\times$ &$\times$ &$\times$ &$\times$ &$\times$ &$\times$ &$\times$ &$\checkmark$ &$\checkmark$ & \\
\hline
\cite{lu2017hybrid} & $\checkmark$ &$\times$& $\times$ &$\times$ &$\times$ &$\times$ &$\times$ &$\times$ &$\times$ &$\times$ &$\times$ &$\times$ &$\times$ & \\
\hline
\cite{alivckovic2017breast} & $\checkmark$ &$\checkmark$ &$\checkmark$ &$\checkmark$ &$\times$ &$\checkmark$ &$\times$ &$\times$ &$\times$ &$\times$ &$\checkmark$ &$\times$ &$\checkmark$ & \\
\hline
\cite{gangavarapu2019novel} &$\checkmark$ &$\times$ &$\times$ &$\times$ &$\times$ &$\times$ &$\times$ &$\times$ &$\times$ &$\times$ &$\times$ &$\times$ &$\checkmark$ &-Feature importance

-chi-square test \\
\hline
\cite{mohammed2018real} &$\checkmark$ &$\checkmark$ &$\times$ &$\checkmark$ &$\times$ &$\checkmark$ &$\times$ &$\checkmark$ &$\checkmark$ &$\times$ &$\times$ &$\times$ &$\times$ & \\
\hline
\cite{khanduzi2019fast} &$\checkmark$ &$\times$ &$\times$ &$\times$ &$\times$ &$\times$ &$\times$ &$\times$ &$\checkmark$ &$\times$ &$\times$ &$\checkmark$ &$\times$ & \\
\hline
\cite{subasi2019epileptic} &$\checkmark$ &$\times$ &$\times$ &$\checkmark$ &$\checkmark$ &$\checkmark$ &$\checkmark$ &$\checkmark$ &$\checkmark$ &$\times$ &$\times$ &$\times$ &$\times$ & Fitness classification accuracy\\
\hline
\cite{soufan2015dwfs} &$\times$ &$\times$ &$\times$ &$\checkmark$ &$\checkmark$ &$\checkmark$ &$\checkmark$ &$\checkmark$ &$\checkmark$ &$\checkmark$ &$\checkmark$ &$\checkmark$ &$\times$ & -Stability

-G-Mean\\
\hline
\cite{li2016genetic} &$\checkmark$ &$\checkmark$ &$\checkmark$ &$\checkmark$ &$\checkmark$ &$\checkmark$ &$\checkmark$ &$\checkmark$ &$\checkmark$ &$\times$ &$\times$ &$\times$ &$\times$ & \\
\hline
\cite{tangherloni2019genhap} &$\checkmark$ &$\times$ &$\times$ &$\times$ &$\times$ &$\times$ &$\times$ &$\times$ &$\times$ &$\times$ &$\times$ &$\checkmark$ &$\times$ & Convergence rate for Average Best Fitness\\
\hline
\cite{moraes2017gass} &$\checkmark$ &$\times$ &$\times$ &$\times$ &$\times$ &$\times$ &$\times$ &$\times$ &$\times$ &$\times$ &$\times$ &$\checkmark$ &$\times$ & \\
\hline
\cite{zhang2019selection}&$\times$ &$\checkmark$ &$\checkmark$ &$\times$ &$\times$ &$\times$ &$\times$ &$\times$ &$\times$ &$\times$ &$\times$ &$\times$ &$\times$ & \\
\hline
\cite{ramadan2016protein} &$\times$ &$\times$ &$\times$ &$\checkmark$ &$\checkmark$ &$\checkmark$ &$\times$ &$\times$ &$\checkmark$ &$\checkmark$ &$\checkmark$ &$\times$ &$\times$ & Discard Ratio\\
\hline
\cite{sayed2019nested} &$\checkmark$ &$\times$ &$\times$ &$\checkmark$ &$\checkmark$ &$\checkmark$ &$\checkmark$ &$\times$ &$\times$ &$\times$ &$\times$ &$\times$ &$\times$ & \\
\hline
\cite{lee2018comprehensive} &$\times$ &$\times$ &$\times$ &$\checkmark$ &$\times$ &$\checkmark$ &$\checkmark$ &$\times$ &$\checkmark$ &$\checkmark$ &$\checkmark$ &$\times$ &$\times$ & \\
\hline
\cite{oztekin2018decision} &$\checkmark$ &$\times$ &$\times$ &$\checkmark$ &$\checkmark$ &$\checkmark$ &$\checkmark$ &$\checkmark$ &$\checkmark$ &$\checkmark$ &$\checkmark$ &$\times$ &$\times$ & G-Mean\\
\hline
\cite{corus2017standard} &$\times$ &$\times$ &$\times$ &$\times$ &$\times$ &$\times$ &$\times$ &$\times$ &$\times$ &$\times$ &$\times$ &$\checkmark$ &$\times$ & \\
\hline
\cite{hashem2017comparison} &$\checkmark$ &$\checkmark$ &$\checkmark$ &$\checkmark$ &$\checkmark$ &$\checkmark$ &$\checkmark$ &$\checkmark$ &$\checkmark$ &$\times$ &$\times$ &$\times$ &$\times$ & \\
\hline
\cite{ansotegui2015model} &$\checkmark$ &$\times$ &$\times$ &$\times$ &$\times$ &$\times$ &$\times$ &$\times$ &$\times$ &$\times$ &$\times$ &$\times$ &$\times$ & \\
\hline
\cite{ghosh2016incorporating} &$\times$ &$\times$ &$\times$ &$\times$ &$\times$ &$\times$ &$\times$ &$\times$ &$\times$ &$\times$ &$\times$ &$\times$ &$\times$ & Dice Similarity\\
\hline
\cite{hemanth2019modified} &$\checkmark$ &$\checkmark$ &$\checkmark$ &$\checkmark$ &$\times$ &$\times$ &$\times$ &$\checkmark$ &$\checkmark$ &$\times$ &$\times$ &$\times$ &$\times$ & \\
\hline
\cite{bhardwaj2016novel} &$\checkmark$ &$\checkmark$ &$\checkmark$ &$\checkmark$ &$\times$ &$\times$ &$\times$ &$\checkmark$ &$\checkmark$ &$\times$ &$\times$ &$\times$ &$\times$ & \\
\hline
\cite{tan2016genetic} &$\checkmark$ &$\checkmark$ &$\checkmark$ &$\times$ &$\times$ &$\times$ &$\times$ &$\times$ &$\times$ &$\times$ &$\times$ &$\times$ &$\times$ & \\
\hline
\cite{tan2018genetic} &$\checkmark$ &$\checkmark$ &$\checkmark$ &$\times$ &$\times$ &$\times$ &$\times$ &$\times$ &$\times$ &$\times$ &$\times$ &$\times$ &$\times$ & \\
\hline
\cite{dashtban2017gene} &$\checkmark$ &$\times$ &$\times$ &$\times$ &$\times$ &$\times$ &$\times$ &$\times$ &$\times$ &$\times$ &$\times$ &$\checkmark$ & $\checkmark$& Laplacian-score, Fisher-score \\
\hline
\cite{liu2019deep} &$\checkmark$ &$\times$ &$\times$ &$\times$ &$\times$&$\times$&$\times$&$\times$&$\times$&$\times$&$\times$&$\times$&$\times$ &  \\
\hline
\cite{fogue2016non} &$\times$ &$\times$ &$\times$ &$\times$ &$\times$ &$\times$ &$\times$ &$\times$ &$\times$ &$\times$ &$\times$ &$\times$ &$\times$ & -Ambulance Usage 

-Patient Waiting time\\
\hline
\cite{plawiak2018novel} &$\checkmark$ &$\times$ &$\times$ &$\checkmark$ &$\checkmark$ &$\checkmark$ &$\checkmark$ &$\checkmark$ &$\checkmark$ &$\times$ &$\times$ &$\checkmark$ &$\checkmark$ & -Sum of Errors

-k-coefficient

-Acceptance feature coefficient\\
\hline
\cite{la2019multidimensional} & $\checkmark$& $\checkmark$& $\checkmark$&$\checkmark$ & $\checkmark$& $\checkmark$& $\checkmark$& $\checkmark$& $\checkmark$& $\checkmark$& $\checkmark$& $\times$& $\times$& \\
\hline
\cite{devarriya2020unbalanced} & $\checkmark$& $\checkmark$& $\checkmark$& $\checkmark$& $\checkmark$& $\checkmark$& $\checkmark$& $\checkmark$& $\checkmark$& $\checkmark$& $\checkmark$& $\times$& $\times$& \\
\hline
\end{longtable}
\end{center}




\end{backmatter}
\end{document}